\documentclass[wcp]{jmlr}

\usepackage{longtable}

\usepackage{booktabs}

\usepackage{graphicx}
\usepackage{amsmath}
\usepackage{amssymb}
\usepackage{booktabs}
\usepackage{multirow}
\usepackage{caption}
\usepackage{subcaption}

\usepackage{arydshln}

\makeatletter
\let\Ginclude@graphics\@org@Ginclude@graphics 
\makeatother

\jmlrvolume{}
\jmlryear{}
\jmlrworkshop{}

\title[Top-K Logits Integration]{Out-of-Distribution Detection with Adaptive Top-K Logits Integration}

 \author{\Name{Hikaru Shijo} \Email{shijo.hikaru@jp.panasonic.com}\\
  \Name{Yutaka Yoshihama} \Email{yoshihama.yutaka@jp.panasonic.com}\\
  \Name{Kenichi Yadani} \Email{yadani.kenichi@jp.panasonic.com}\\
  \Name{Norifumi Murata} \Email{murata.norifumi@jp.panasonic.com}\\
  \addr Panasonic Automotive Systems Co., Ltd.}

\begin{document}

\maketitle

\begin{abstract}
    Neural networks often make overconfident predictions from out-of-distribution (OOD) samples. Detection of OOD data is therefore crucial to improve the safety of machine learning. The simplest and most powerful method for OOD detection is MaxLogit, which uses the model’s maximum logit to provide an OOD score. We have discovered that, in addition to the maximum logit, some other logits are also useful for OOD detection. Based on this finding, we propose a new method called ATLI (Adaptive Top-k Logits Integration), which adaptively determines effective top-k logits that are specific to each model and combines the maximum logit with the other top-k logits. In this study we evaluate our proposed method using ImageNet-1K benchmark. Extensive experiments showed our proposed method to reduce the false positive rate (FPR95) by 6.73\% compared to the MaxLogit approach, and decreased FPR95 by an additional 2.67\% compared to other state-of-the-art methods.
\end{abstract}

\section{Introduction}

\label{sec:intro}

\begin{figure}[t]
  \centering
  \def\subfigcapskip{-5pt}
  \begin{minipage}[c]{0.49\textwidth}
    \centering
    \subfigure[Overview]{%
      \includegraphics[width=\textwidth]{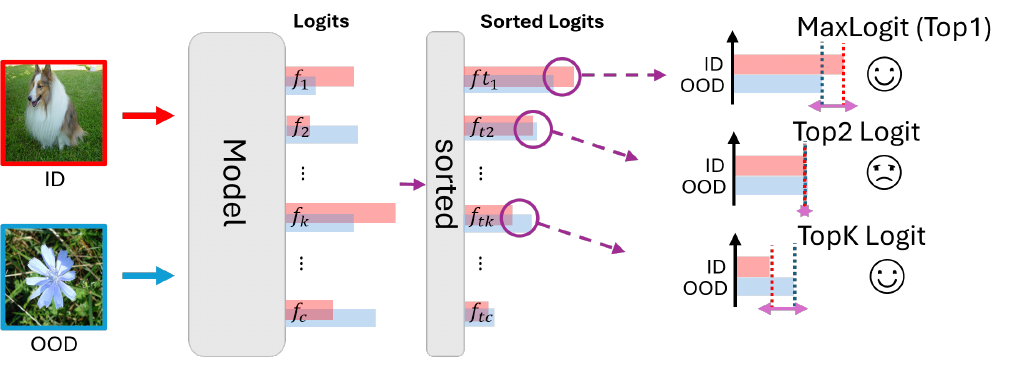}
    }
  \end{minipage}%
  \hfill
  \begin{minipage}[c]{0.49\textwidth}
    \centering
    \subfigure[Logit distributions on Eva \citep{eva}]{%
      \includegraphics[width=\textwidth]{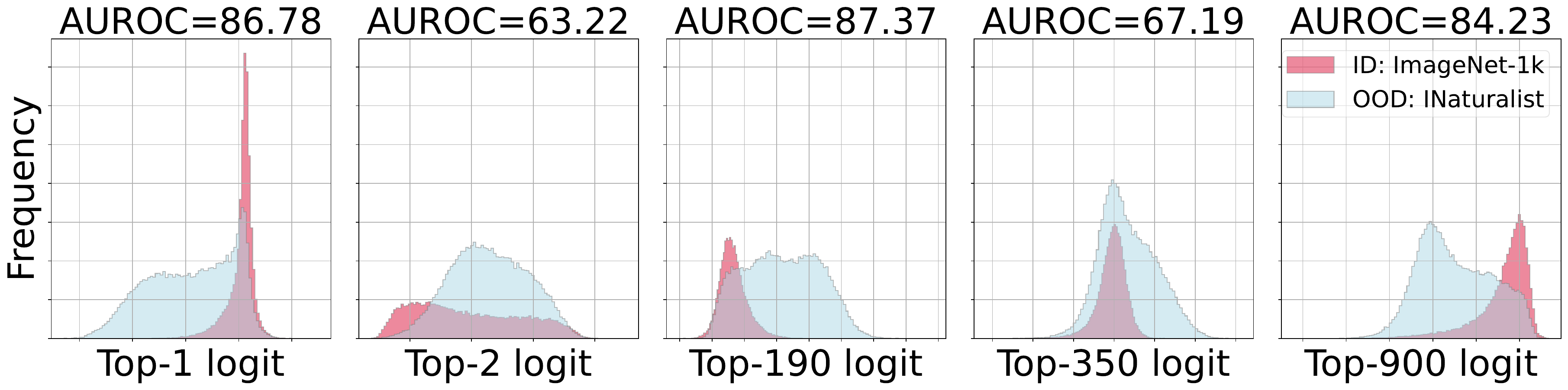}
    }\\
    \subfigure[Logit distributions on ResNet-50d \citep{resnet50d2019}]{%
      \includegraphics[width=\textwidth]{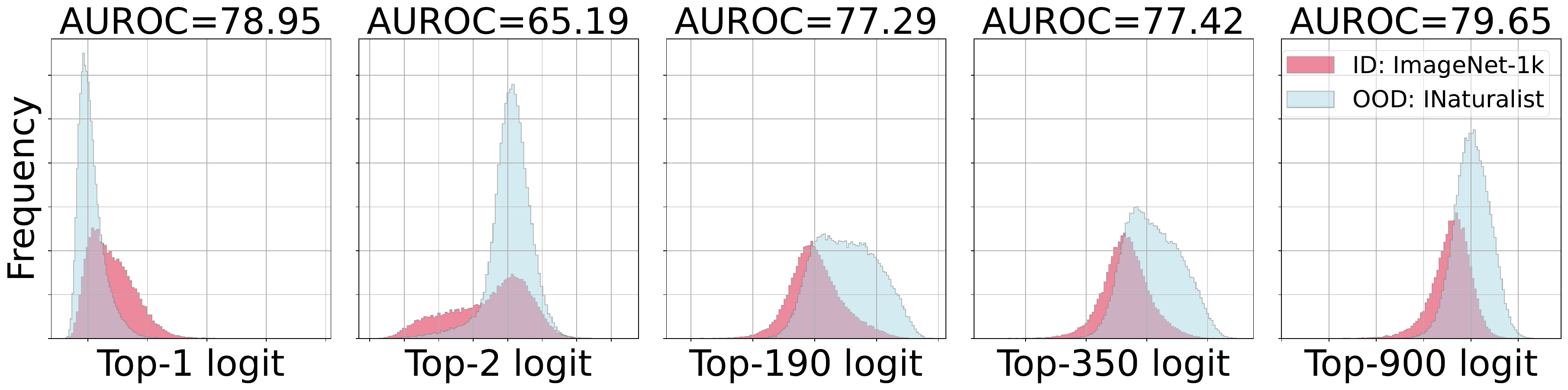}
    }
  \end{minipage}
  \caption{Illustration of our idea and the actual distribution of model logits. (a) is a conceptual diagram for calculating logits, illustrating that top-k logits other than MaxLogit can separate ID and OOD. Please also note that logits are sorted for each sample. (b) and (c) show the distribution of top-k logits for ID and OOD in the actual model. These models were trained using ImageNet-1K as the ID dataset. INaturalist is used as an OOD dataset to measure the performance of OOD detection. The red area represents ImageNet-1K and the blue area represents INaturalist.}
  \label{fig:compare_dist_class_logit}
  \end{figure}

Out-of-distribution (OOD) detection is a critical task for improving the reliability and safety of machine learning models, particularly in real-world applications such as autonomous driving and medical diagnostics. This is because models tend to make overly confident predictions when faced with data that diverges from their training distribution. Several existing methods for OOD detection are commonly employed, including Maximum Softmax Probability (MSP) \citep{msp17iclr}, MaxLogit \citep{maxlogit}, and Energy \citep{energyood20nips}. MSP uses maximum softmax probability, MaxLogit uses maximum logits, and Energy uses the logsumexp function for logits. These methods compute scores based on logit space.
MaxLogit considers only maximum logits, whereas MSP and Energy consider all class logits. These methods are simple and powerful, but they are not state-of-the-art. However, since logits contain high-level semantic information, there is still potential for their use. In this study, we discovered that considering all logits or only maximum logits can result in low scores. In Figure \ref{fig:compare_dist_class_logit} (b) and (c), we show the distribution of in-distribution (ID) and OOD with different top-k logits. The leftmost figure shows the distribution of the top-1 logit (MaxLogit), which is typically used in OOD detection. The other figures show the distributions of different top-k logits. In Eva (b), the top-2 and top-350 logits largely overlap between ID and OOD. However, the top-190 and top-900 logits separate ID and OOD as effectively as MaxLogit. In ResNet-50d (c), however, the top-2 largely overlap between ID and OOD, whereas the other top-190, top-350, and top-900 separate ID and OOD as effectively as MaxLogit. This pattern is also observed in other models. We thus find that the top-k logits effective for OOD detection vary across models. These findings suggest that the OOD score should contain only the effective top-k logits in each model.

In this paper, we propose ATLI, which adaptively selects a set of effective top-k logits for each trained model, excluding the top-1 logit, and combines them with the maximum logit to compute the OOD score. To select effective top-k logits, we use a pseudo-OOD sample created from a training sample and evaluated our proposed method using various models and datasets. The results of the experiments reveal that the latest methods depend on the model, and in cases with poorly compatible models, the accuracy is below that of MaxLogit, which we use as our baseline. However, our proposed method consistently outperforms existing baselines across a variety of trained models, indicating its low dependency on specific trained models.
Our contributions are summarized as follows.
\begin{list}{\textbullet}{\topsep=0pt}
  \setlength{\itemsep}{0cm}
  \setlength{\parskip}{0cm}
    \item We reveal that there are several top-k logits that can separate ID and OOD as effectively as MaxLogit for each trained model.
    \item We develop a methodology for identifying effective top-k logits for each trained model by utilizing pseudo-OOD.
    \item Our proposed method has a very simple implementation and outperforms other methods.
\end{list}

\section{Related Work}
\label{sec:related}
\subsection{Score Design method}
The most basic approach to OOD detection is to design a scoring function that can separate ID and OOD based on the output of a pre-trained neural network model.
Hendrycks et al. \citep{msp17iclr} adopt a simple baseline using the maximum softmax probability. Similarly, MaxLogit \citep{maxlogit} uses the maximum value of the predicted logits as the score. 
The energy score \citep{energyood20nips} computes the logsumexp of logits. These methods are based on logits or the probability of neural networks.
On the other hand, some studies use the features of the penultimate layer for OOD detection.
Lee et al. \citep{lee2018simpleunifiedframeworkdetecting} use the Mahalanobis distance, which computes the distance of class-wise Gaussian distributions on training data. Sun et al. \citep{sun2022outofdistributiondetectiondeepnearest} use the KNN method for OOD detection.
In recent years, Wang et al. \citep{vim2022cvpr} have proposed ViM, which is both logit-based and feature-based. In another straightforward approach, Yu et al. \citep{featurenorm2023cvpr} selected the valid layer for OOD detection using a feature ratio. More recently, GEN \citep{gen2023cvpr} uses the top 10 percentile of sorted probabilities. TRIM \citep{trim2024} uses the top-7 to top-16 sorted probabilities of models for OOD detection. However, softmax normalization compresses the logit distribution, especially when logits are close in magnitude. This obscures fine-grained differences among classes and may result in the loss of valuable semantic cues needed for OOD detection. Therefore we use the region of logits which contains high-level semantic information. \\

\subsection{Training method}
Another approach to OOD detection is to focus on training model by OOD sample.
Hendrycks et al. \citep{OE} tackle this by re-training using a new loss function that incorporates class label loss and out-of-distribution loss.
As follow-up work, OECC \citep{OECC} were able to suppress excessive confidence in the model by adding a loss term that aligns confidence for in-distribution training samples with training accuracy.
However, these training methods require real OOD data. VOS \citep{vosiclr2022} uses pseudo-OOD sampled from the low-likelihood region of the class-conditional distribution. NPOS \citep{nposiclr2023} use non-parametric outlier synthesis, which does not make any distributional assumptions as to the ID embeddings. The use of a training method is the most effective strategy for OOD detection; however, it requires time-consuming procedures such as model re-training. And training with additional OOD datasets may negatively affect the model's accuracy.
\subsection{Enhancement methods}
Some studies focus on improving the accuracy of OOD detection using designed scores such as MSP and energy by adding constraints to the model’s intermediate representations or inputs. ODIN \citep{odin18iclr} enhances MSP scores by adding small perturbations to the input. Sun et al. \citep{react21nips} discovered that internal activation of neural networks results in highly distinctive signature patterns of OOD, and adopted ReAct, which applies feature clipping to the penultimate layer of neural networks. DICE \citep{dice2022eccv} computes the contribution matrix of the product of features and weights by training data, and prunes the weights that are below a certain threshold based on this contribution.
ASH \citep{ash2022} prunes the activations of final linear layer based on the ratio of the activation.
However, because these methods cause changes in weights or features, they affect the accuracy of the model.
In recent work, LTS \citep{lts2024} and Scale \citep{scale2024} enhance energy scores without affecting the model’s accuracy, by scaling logits based on the top percentage of activations.

\section{Preliminaries}
We consider here a neural network model for $C$-class image classification. In general, a neural network represents a mapping function $f: X \to Y$, where $X$ is the input space and $Y$ is the target space. Given an input image $\boldsymbol{x} \in \mathbb{R}^{3 \times W \times H}$ that belongs to class $k$, the neural network $f(\boldsymbol{x})$ transforms $x$ into $C$ real-valued numbers known as logits, which are then used to predict the label of the image. 
The ultimate goal of OOD detection is to determine whether the input image $x$
is ID or OOD. Typically, OOD detection is defined by the following discriminative function.
\begin{equation}
G_\lambda(\boldsymbol{x}) = \begin{cases}
    \text{ID} & S(\boldsymbol{x}) \geq \lambda \\
    \text{OOD} & S(\boldsymbol{x}) < \lambda
\end{cases}
\label{eq:defined_score}
\end{equation}
$S(x)$ is a scoring function such as MSP or MaxLogit. By adopting a certain threshold $\lambda$, it becomes possible to distinguish between OOD and ID.

\begin{figure}[t]
  \centering
  \begin{center}
    \begin{minipage}[b]{0.49\columnwidth}
        \centering
        \includegraphics[width=1\columnwidth]{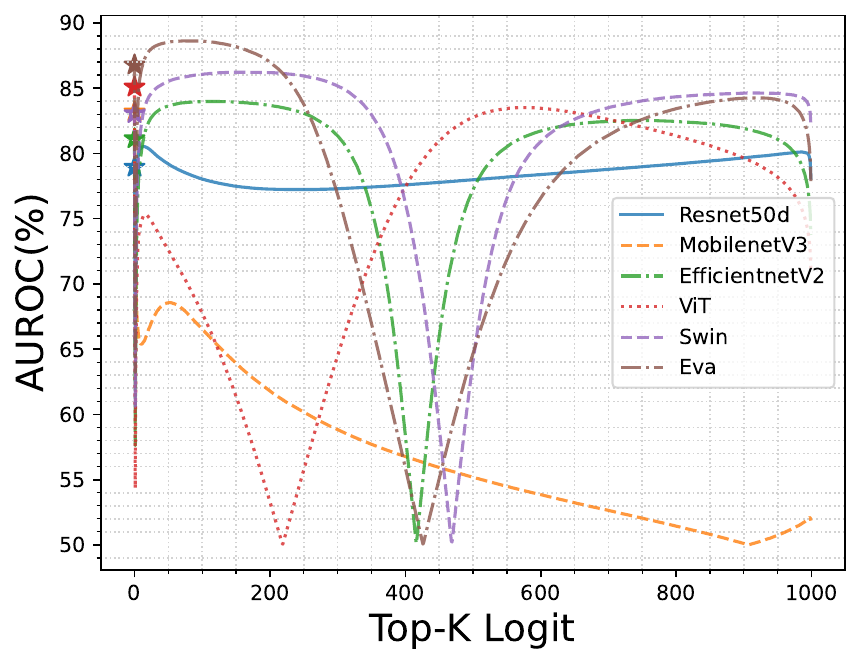}
    \end{minipage}
    \begin{minipage}[b]{0.49\columnwidth}
        \centering
        \includegraphics[width=1\columnwidth]{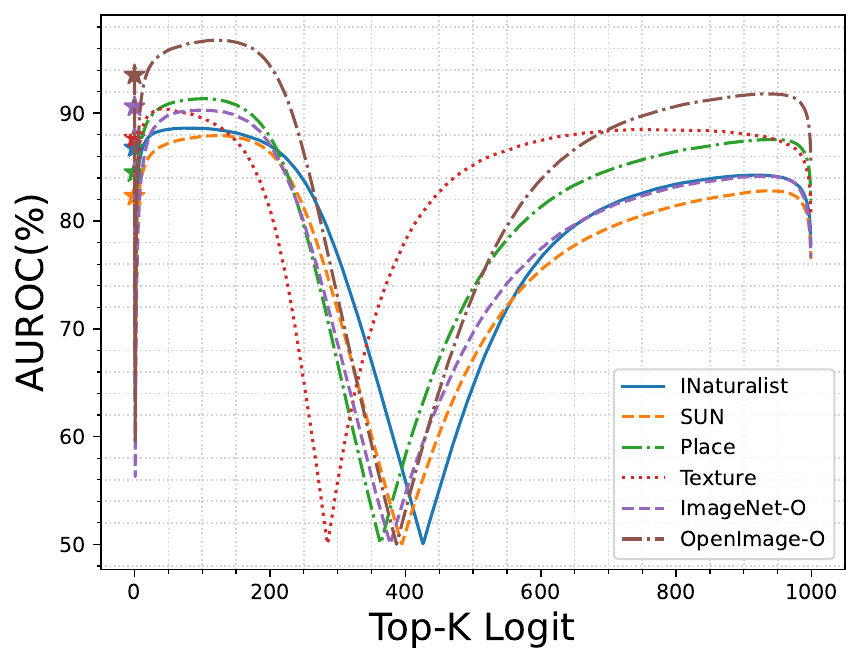}
    \end{minipage}
  \vspace*{-0.6cm}
\end{center}
  \caption{AUROC scores for each top-k. The left chart shows the scores for top-k logits across different models. The OOD data used is INaturalist. The right-hand chart shows the scores for top-k logits for different OOD datasets in Eva.}
  \label{fig:auroc_all_topk}
  \end{figure}
  
\section{Method}
\subsection{Analysis of top-k logit on various models}
We start by observing the top-k logits of various models. Figure \ref{fig:auroc_all_topk} shows the AUROC of top-k logits for eight different models. We observe that all models have scores equal to or greater than MaxLogit (top-1) for a certain top-k logit. On the other hand, there are some top-k logit areas for which the scores are very low (e.g., around top-200 in Swin). We also note that which top-k logits have high scores vary according to the model. The AUROC for different OOD data is shown in the right-hand chart in Figure \ref{fig:auroc_all_topk}. It can be seen that the trends are similar across all types of OOD. From these observations, it is necessary to vary the logits used for each model.

\subsection{Rethinking MaxLogit vs. Energy vs. MSP}
In this section, we reconsider the differences among the three basic methods: MaxLogit, Energy and MSP. \\
The energy score is given by the following formula. 
\begin{equation}
\label{eq: energy_1}
 E(\mathbf{x}; f) = T \cdot \log \sum_{i}^{C}e^{f_i(\mathbf{x}) / T}
\end{equation}
Assume $T=1$ and that the logit takes a maximum value at $i=j$.
Since monotonically increasing functions such as $\exp(\cdot)$ do not affect OOD detection, Eq. \ref{eq: energy_1} can be transformed as follows.
\begin{equation}
\label{eq: energy_2}
  e^{E(\mathbf{x}; f)} = e^{f_j(\boldsymbol{x})} + \sum_{i\ne\ j}^{C}e^{f_i(\boldsymbol{x})}
\end{equation}
The first term of Eq. \ref{eq: energy_2} represents MaxLogit, 
while the second term represents the sum of the exponential of the other logits. 
It can be understood that the second term is relevant to the difference between MaxLogit and the energy score. MSP score is given by the following formula. 
\begin{equation}
\label{eq: msp_1}
MSP(\mathbf{x}; f) = \frac{e^{f_j(\boldsymbol{x})}}{\sum_{i}^{C} e^{f_i(\boldsymbol{x})}}
\end{equation}
Since monotonically increasing functions such as $\log(\cdot)$ do not affect OOD detection, Eq. \ref{eq: msp_1} can be transformed as follows.
\begin{equation}
\label{eq: msp_2}
\log (MSP(\mathbf{x}; f)) = f_j(\boldsymbol{x}) - \log\sum_{i}^{C} e^{f_i(\boldsymbol{x})}
\end{equation}

\begin{table}[t]
    \centering
    \small
    \begin{tabular}{@{}lccc@{}}
        \toprule
        Methods & ResNet-50d & MobileNetV3 & Swin \\ \midrule
        MaxLogit & 80.26 & 66.39 & \textbf{47.94}  \\
        MSP  & \textbf{77.82} & 74.95 & 54.19 \\
        Energy  & 86.06  & \textbf{64.33} & 48.67 \\ \bottomrule
    \end{tabular}
    \caption{OOD detection scores (FPR95) using different methods and models. These models are finetuned in ImageNet-1k. Lower FPR95 values indicate better performance. Swin is an abbreviation of Swin Transformer.
    The best score for each method across the models appear in bold.}
    \label{tab:compare_method}
\end{table}
\noindent
Similar to Energy, for MSP, the first term of Eq. \ref{eq: msp_2} represents MaxLogit, while the second term denotes logsumexp of the logits for all classes, including MaxLogit. It can be understood that the second term is the difference between MaxLogit and MSP.
A simple difference between MSP and Energy lies in the sign of the second term. Energy is positive, whereas MSP is negative.
The differences between the three methods therefore rest on two points. The first uses only one logit or all logits. The second uses signs for logits other than the top-1.
Table \ref{tab:compare_method} represents the FPR95 of MaxLogit, MSP and Energy across the three models: the differences shown in Eq. \ref{eq: energy_2} and \ref{eq: msp_2} are reflected in the differences in scores. 
ResNet-50d has the lowest score for MSP, followed by MaxLogit with the next lowest score. However, in MobileNetV3, Energy is the lowest, followed by MaxLogit, whereas MSP shows a significant increase in score.
Furthermore, in Swin Transformer, MaxLogit has the lowest score, followed by Energy with the next lowest score.
Based on the above, in ResNet-50d, a negative second term is suitable for OOD detection, whereas in MobileNetV3, a positive second term is more appropriate. Furthermore, Swin Transformer is suitable for using only the top-1 logit. From these results, it is clear that the appropriate number of logits and the appropriate  signs for logits other than the top-1 vary for each model. Further detailed analysis, in conjunction with Figure \ref{fig:compare_dist_class_logit}, reveals that the magnitude relationship between ID and OOD changes for each top-k logit. Scoring function is designed to have high value for ID as defined as Eq. \ref{eq:defined_score}. Therefore, rather than assigning the same sign for all logits, it is necessary to assign the appropriate sign for each model and each top-k logit.
Moreover, traditional methods consider only either a single logit or all logits, but Figure \ref{fig:auroc_all_topk} indicates that there are top-k logits that are not suitable for OOD detection. Therefore, only effective logits should be included in the score function.
With this motivation in mind, we discuss a new scoring function in the next section.

\begin{figure*}[t]
\centering
   \includegraphics[scale=0.6]{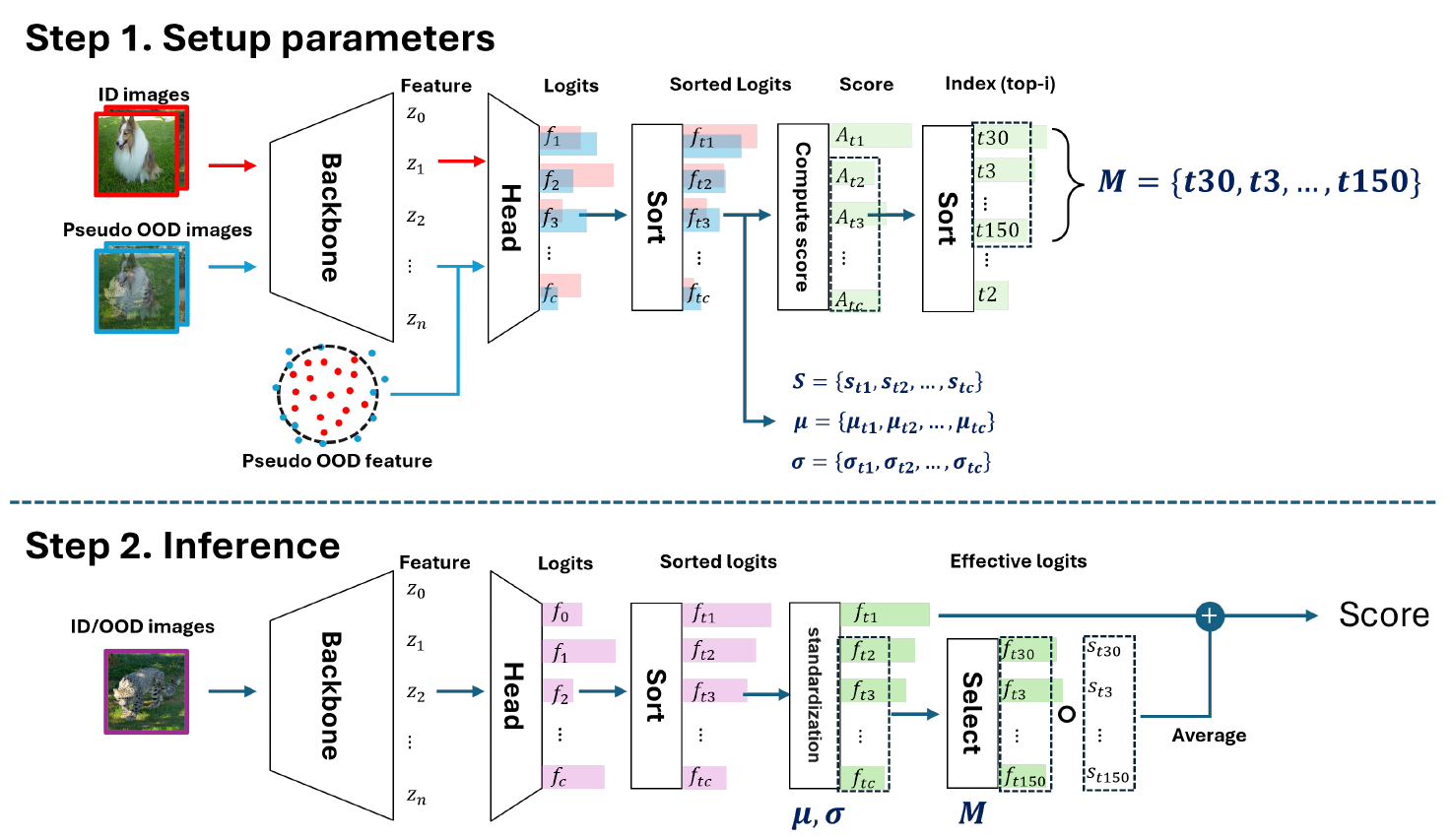}
  \vspace*{-0.3cm}
  \caption{
 An overview of our proposed method, which is divided into two phases. In the first phase, “(1) Setup parameters,” training images and pseudo images are input into the model, logits are sorted, and the scores are computed. The indices of the logits with the top percentage of scores are selected to create a parameter set $M$. In the next phase, “(2) Inference time,” test images are input into the model and a score is computed to determine whether these images are ID or OOD.
 }
  \label{fig:method_overview}
\end{figure*}

\subsection{Our proposed method}
An overview of our proposed method is shown in Figure \ref{fig:method_overview}. 
We aim to include only the top-k logits that are effective for OOD detection in the scoring function.
Our scoring function is the following formula.
\begin{equation}
\label{eq: proposal_method}
    \text{ATLI}(\mathbf{x}; f) = f^{\prime}_{\text{top-}1}(\mathbf{x}) + \frac{1}{|M|}\sum_{i\in M} s_i\cdot f_{\text{top-}i}^{\prime}(\mathbf{x})
\end{equation}

Here, $f^{\prime}_{\text{top-}i}(\mathbf{x})$ is the standardized version of the i-th largest logit $f_{\text{top-}i}(\mathbf{x})$, defined as
$
f^{\prime}_{\text{top-}i}(\mathbf{x}) = (f_{\text{top-}i}(\mathbf{x}) - \mu_i)/ \sigma_i
$
, where the mean $\mu_i$ and standard deviation $\sigma_i$ are computed from the top-i logits of the training data. This standardization ensures that all top-k logits are normalized and can be treated equally in the scoring function.  
The first term of the Eq. \ref{eq: proposal_method}, $f^{\prime}_{\text{top-}1}(\mathbf{x})$, represents MaxLogit, which corresponds to the largest logit. The second term represents other top-k logits that are effective for OOD detection. $M$ is the set of indices of the top-k logits effective for OOD detection, excluding the top-1. $s_i$ is a parameter that assigns a sign to each top-k logit. The $|M|$ represents the number of elements in $M$.
The details of how the set $M$ and the parameters $s_i$ are determined are explained in the following paragraph.

\paragraph{Determining $M$}
\label{seq:determin_M}
This section describes how to determine a subset $M$ of logit's indices that are effective for OOD detection. We decide set $M$ using a pseudo-OOD. This method is illustrated in Figure \ref{fig:method_overview} (above). The first step is to prepare a pseudo-OOD which is generated from ID training samples. The specifics of this pseudo-OOD are detailed in Section \ref{sec:pseudo_ood_desing}. 
To obtain logits, the models draw inferences from both the training data and the pseudo-OOD data. We compute an OOD score defined as $Score = AUROC - FPR95$ for each top-k logit. The top-k logit with high scores for the pseudo-OOD also appear to be effective for OOD in the test environment. The indices of the logits in the top few percent of scores are therefore selected as valid logits and designated as set $M$. Note that the top-1 is not included here.

\paragraph{Determining sign}
The scoring function is designed to yield high values for ID samples, as defined in Eq. \ref{eq:defined_score}. However, as shown in Figure \ref{fig:compare_dist_class_logit} (b) and (c), we confirmed that, except for the top-1 logit, the magnitude relationship of top-k logit between ID and OOD varies according to the model. We therefore determine the signs for the top-k by using pseudo-OOD to observe the tendencies of the model. The sign is determined such that the distribution of the top-k logits with the sign for ID is always higher than that with the sign of the OOD.
Let $\mu_i$ be the average of all top-i logits inferred by the model from the entire set of training data $\mathbf{X}=\{\boldsymbol{x}_1, ..., \boldsymbol{x}_D\}$. This can be expressed as $\mu_i= \frac{1}{D}\sum_{n=1}^D f_{top-i}(\boldsymbol{x}_n)$. Conversely, let $\mu^{\prime}_i$ be the average of all top-i logits inferred by the model from the pseudo-OOD, which is defined in the same way. We compute the sign following this formula.
\begin{equation}
s_i = \begin{cases}
    1 & \mu_i \geq \mu_i^{\prime} \\
    -1 & \mu_i < \mu_i^{\prime}
\end{cases}
\label{eq:sign}
\end{equation}

\subsection{Pseudo OOD design}
\label{sec:pseudo_ood_desing}
We generate pseudo-OOD samples using a combination of Mixup \citep{mixupiclr2018} and VOS \citep{vosiclr2022}. Mixup creates convex combinations of two images. Typically, Mixup uses a mixing ratio sampled from a beta distribution. However, we set the mixing ratio to 0.5 to ensure that the OOD samples are evenly mixed between the two classes.
However, using Mixup alone does not adequately cover the input space, as it only generates OOD samples that remain within the ID distribution. For effective pseudo-OOD generation, it is crucial to ensure diversity that spans a larger input space \citep{cnc2022}.
To address this, in the penultimate layer’s feature, we generate OOD samples outside the ID distribution using a single Gaussian distribution that encompasses all of the training data. Here, it is important to note that the conventional VOS assumes a Gaussian distribution for each class, which distinguishes our method from this approach.
Since Mixup generates samples within the ID data and VOS generates samples outside the ID data, combining these methods allows us to cover a much wider feature space.
To summarize the above, the convex combination $\bar{x}$ of the two training images $x_a$ and $x_b$ from different classes in training data can be represented as
$
\bar{\boldsymbol{x}} = 0.5\boldsymbol{x}_a + 0.5 \boldsymbol{x}_b
$.
Additionally, by assuming that the penultimate feature's training data follows a single Gaussian distribution, the features are sampled from its low likelihood;
$
    \mathbf{z^{\prime}} \sim \mathcal{N}(\hat{\mathbf{\mu}}, \hat{\mathbf{\Sigma}})
$.
Here, $\hat{\mathbf{\mu}}$ represents the mean of the Gaussian distribution calculated from the training data, while $\hat{\mathbf{\Sigma}}$ denotes its covariance matrix, also derived from the training data.
The items generated from the two equations above are combined in a 1:1 ratio to create the pseudo-OOD.

\begin{figure*}[t]
  \centering
  \includegraphics[scale=0.2]{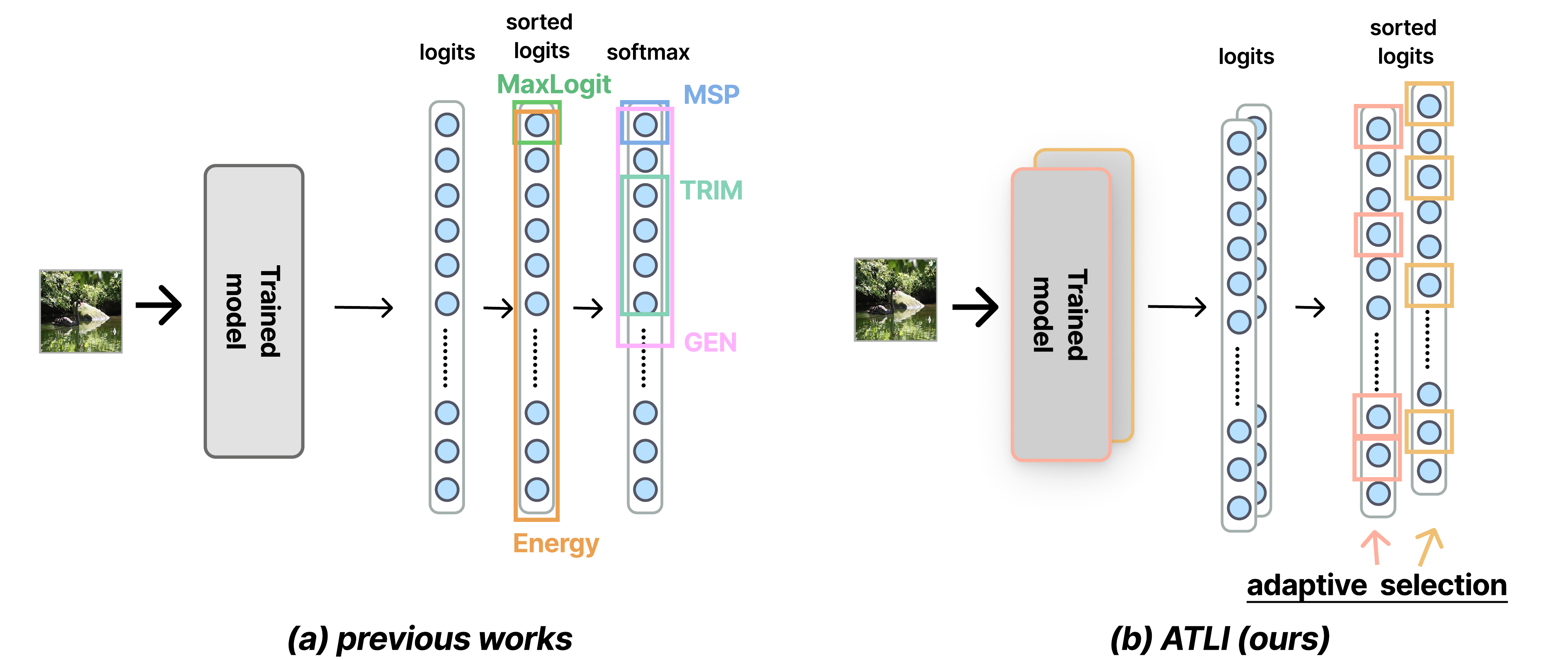}
  \caption{Comparison between prior works and our proposed ATLI. (a) Prior methods use fixed selection strategies. For example, MaxLogit uses only the maximum logit, and TRIM uses the top-6 to top-15 softmax probabilities. These approaches are model-agnostic.
(b) Our method, ATLI, adaptively selects a small subset of top-k logits for each model, enabling more effective OOD detection based on the characteristics of each model.}
  \label{fig:diff__previous_work}
\end{figure*}

\section{Experiments}

\subsection{Experimental Settings}
\textbf{Datasets}
\label{seq:datasets}
We evaluated OOD detection using ImageNet-1K benchmark. For ImageNet-1K benchmark, 
we evaluated OOD detection on ImageNet-1K \citep{imagenet} as ID. As OOD datasets, we use INaturalist \citep{inaturalist2018}, SUN \citep{sunieee2010}, Place \citep{placeieee2018}, Texture \citep{textureieee2014}, ImageNet-O \citep{imagenet-ocvpr2021} and OpenImage-O \citep{vim2022cvpr}. ImageNet-1K is a 1000-classification task, making this benchmark suitable for our proposed method, which uses certain numbers of logits.  \\\\

\begin{table}[h]
    \centering
    \begin{tabular}{@{}lcc@{}}
        \toprule
        Model & Acc(\%) & Parameters(M)\\ \midrule
        ResNet-50d &77.22 &25.6 \\
        MobileNetV3 &77.90 &5.5 \\
        EfficientNetV2 &84.77 & 54.1 \\
        Vision Transformer (Vit) & 88.17 &  304.2 \\
        Swin Transformer (Swin) & 85.27 & 87.8 \\
        Eva & 88.59 & 304.1 \\
        \bottomrule
    \end{tabular}
    \caption{Details of the models in ImageNet-1K benchmark.}
    \label{tab:model_detail_imagenet}
\end{table}

\noindent \textbf{Models}
In ImageNet-1K benchmark, we used various models that incorporated CNN and Transformer-based architectures. The CNN-based architecture includes ResNet-50d \citep{resnet50d2019}, MobileNetV3 \citep{mobilenetv3} and EfficientNet \citep{effnetv2}.
The Transformer-based architecture includes Vision Transformer \citep{vit}, Swin Transformer \citep{swin2021iccv}, and Eva02 \citep{eva}. For the ImageNet-1K benchmark, we used pre-trained weights in timm \citep{rw2019timm}. Detailed information on these models is provided in Table \ref{tab:model_detail_imagenet}. \\\\
\textbf{Evaluation metrics}
We used two commonly used metrics for OOD detection. AUROC shows to what degree the ID and OOD distributions are separated, with higher scores indicating better separation. FPR95 represents the false positive rate when the true positive rate is 95\%. A lower value indicates a better score. \\\\
\textbf{Post-hoc methods}
We conducted a comparison with existing methods to evaluate our proposed approach. MSP, MaxLogit and Energy are a fundamental baseline for evaluating our method, with ReAct, Dice and Scale enhancing the OOD method. Finally we added ViM, GEN and TRIM, which are strong OOD methods. ViM is a combination of logit and feature-based strategies, and GEN and TRIM are probability-based regional methods. The difference between the baseline methods and ours is visualized in Figure \ref{fig:diff__previous_work}.\\\\
\textbf{Implementation details}
On ImageNet-1K benchmark, when estimating the parameter of each trained model, we sampled $D = 100,000$ images randomly sampled from the entire set of training data.

\begin{table*}[t]
  \centering
  \small
  \resizebox{\textwidth}{!}{
  \begin{tabular}{@{}l c@{}lc@{}lc@{}lc@{}lc@{}lc@{}lc@{}l @{}}
    \toprule \\
    \multirow{2}{*}{\textbf{Method}} & \multicolumn{2}{c}{\textbf{ResNet-50d}} & \multicolumn{2}{c}{\textbf{MobileNetv3}} & \multicolumn{2}{c}{\textbf{EffienetNetV2}} & \multicolumn{2}{c}{\textbf{Vit}}  & \multicolumn{2}{c}{\textbf{Swin}} & \multicolumn{2}{c}{\textbf{Eva}} & \multicolumn{2}{c}{\textbf{Average}}\\
    & \scriptsize AUROC\(\uparrow\) & \scriptsize FPR95\(\downarrow\)& \scriptsize AUROC\(\uparrow\) & \scriptsize FPR95\(\downarrow\)& \scriptsize AUROC\(\uparrow\) & \scriptsize FPR95\(\downarrow\)& \scriptsize AUROC\(\uparrow\) & \scriptsize FPR95\(\downarrow\) & \scriptsize AUROC\(\uparrow\) & \scriptsize FPR95\(\downarrow\) & \scriptsize AUROC\(\uparrow\) & \scriptsize FPR95\(\downarrow\) & \scriptsize AUROC\(\uparrow\) & \scriptsize FPR95\(\downarrow\) \\
    \midrule \\
    MSP &75.70	&77.82	&77.64	&74.95	&81.87	&56.76	&85.85	&45.21	&83.68	&54.19	&87.98	&40.88	&82.12	&58.30 \\
    MaxLogit &74.76	&80.26	&83.17	&66.39	&79.06	&55.93	&83.21	&39.92	&83.56	&47.94	&87.58	&36.63	&81.89	&54.51 \\
    Energy &73.90	&86.06	&83.61	&64.33	&75.96	&63.7	&81.38	&40.86	&82.18	&48.67	&86.92	&34.78	&80.66	&56.40 \\
    ReAct &73.43	&90.02	&\underline{83.97}	&\underline{63.78}	&74.93	&86.06	&84.08	&37.66	&83.99	&46.34	&89.23	&33.10	&81.61	&59.49 \\
    DICE &72.03	&82.49	&71.51	&85.80	&49.96	&91.59	&71.24	&63.59	&80.34	&48.76	&89.72	&31.77	&72.47	&67.33 \\
    ReAct+DICE &73.03	&80.79	&71.57	&85.90	&45.50	&97.61	&76.64	&56.51	&82.52	&46.54	&90.37	&32.01	&73.27	&66.56 \\
    LTS &73.76	&87.69	&83.42	&\textbf{62.96}	&72.06	&66.01	&82.03	&40.18	&83.31	&47.38	&86.50	&35.09	&80.18	&56.55 \\
    Scale &74.87	&79.48	&80.04	&70.73	&58.33	&90.41	&80.25	&42.34	&83.17	&48.77	&77.48	&43.80	&75.69	&62.59 \\
    ViM &77.85	&77.49	&80.38	&75.71	&\textbf{87.49}	&47.37	&\textbf{92.30}	&35.45	&\textbf{88.98}	&48.60	&\textbf{92.93}	&\underline{29.92}	&\underline{86.66}	&52.42 \\
    TRIM &75.64	&\underline{74.08}	&76.84	&76.78	&71.73	&73.45	&78.35	&49.26	&77.93	&60.99	&82.74	&46.21	&77.21	&63.46 \\
    GEN & \underline{78.09}	&75.41	&81.91	&69.86	&84.58	&\underline{47.42}	&88.92	&\underline{33.56}	&87.05	&\underline{45.56}	&91.40	&30.86	&85.33	&\underline{50.45} \\
    ATLI (Ours) &\textbf{78.68}	&\textbf{71.31}	&\textbf{84.14}	&65.40	&\underline{87.14}	&\textbf{44.27}	&\underline{90.60}	&\textbf{32.31}	&\underline{88.30}	&\textbf{43.97}	&\underline{92.28}	&\textbf{29.43}	&\textbf{86.86}	&\textbf{47.78} \\
    \bottomrule
  \end{tabular}
  }
  \caption{
  OOD detection for our method and the baseline methods. The ID dataset is ImageNet-1K, and the OOD datasets are INaturalist, SUN, Place, Texture, OpenImage-O and ImageNet-O. These results represent the average AUROC and average FPR95 over the six OOD datasets. AUROC and FPR95 are shown as percentages. The best results appear in bold, and the second best are underlined. ATLI uses 10\% of all logits ($|M|=100$), adaptively selected for each model.
    }
    \label{tab:results_all_model}
\end{table*}

\subsection{Results on ImageNet-1K benchmark}
The results are summarized in Table \ref{tab:results_all_model}, where we report AUROC (↑) and FPR95 (↓) for each method.
ATLI consistently outperforms all baselines in both AUROC and FPR95 across nearly all models, achieving the highest average AUROC (86.86\%) and the lowest average FPR95 (47.78\%) among the 12 methods evaluated. 
Compared to MaxLogit, which relies solely on the largest logit, and Energy, which aggregates all logits, ATLI demonstrates a substantial improvement. These results indicate that simply using the maximum logit (as in MaxLogit) or using all logits (as in Energy) is suboptimal. Rather than relying on only the highest logit or including all logits, we find that selecting a small number of the most informative logits in the middle range leads to better OOD detection performance.
In particular, TRIM, which is conceptually close to our approach as it leverages a fixed subset of softmax probabilities (top-6 to top-15), serves as a direct competitor. However, ATLI surpasses TRIM across all metrics: achieving \text{+}9.65 points higher AUROC and \verb|-|15.68 points lower FPR95 on average. 
Moreover, ATLI surpasses recent SOTA methods such as GEN and ViM, which have shown strong performance in prior works. On average, ATLI improves over GEN by \text{+}1.53 AUROC and \verb|-|2.67 FPR95, and over ViM by \text{+}0.20 AUROC and \verb|-|4.64 FPR95.
These results support our hypothesis that incorporating model-adaptive selected top-k logits, determined via pseudo-OOD samples, leads to more effective OOD scoring than relying solely on the maximum logit or fixed heuristics.

\subsection{Additional Results}
\label{sec: additional_results_imagenet-1k}

\begin{figure}[t]
  \centering
  \begin{center}
    \begin{minipage}[b]{0.49\columnwidth}
        \centering
        \includegraphics[width=1\columnwidth]{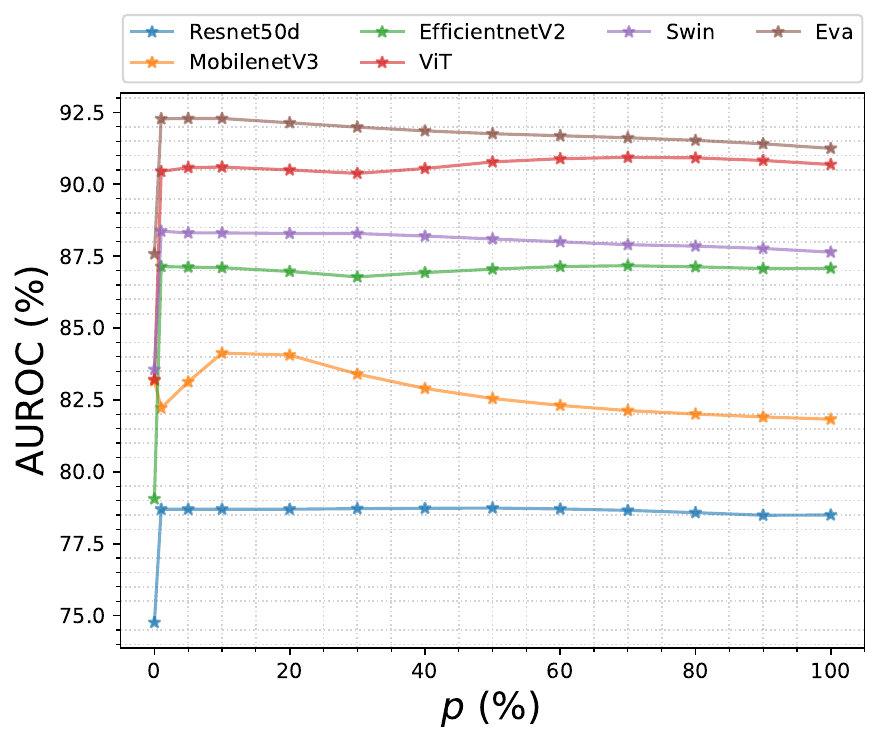}
    \end{minipage}
    \begin{minipage}[b]{0.49\columnwidth}
        \centering
        \includegraphics[width=1\columnwidth]{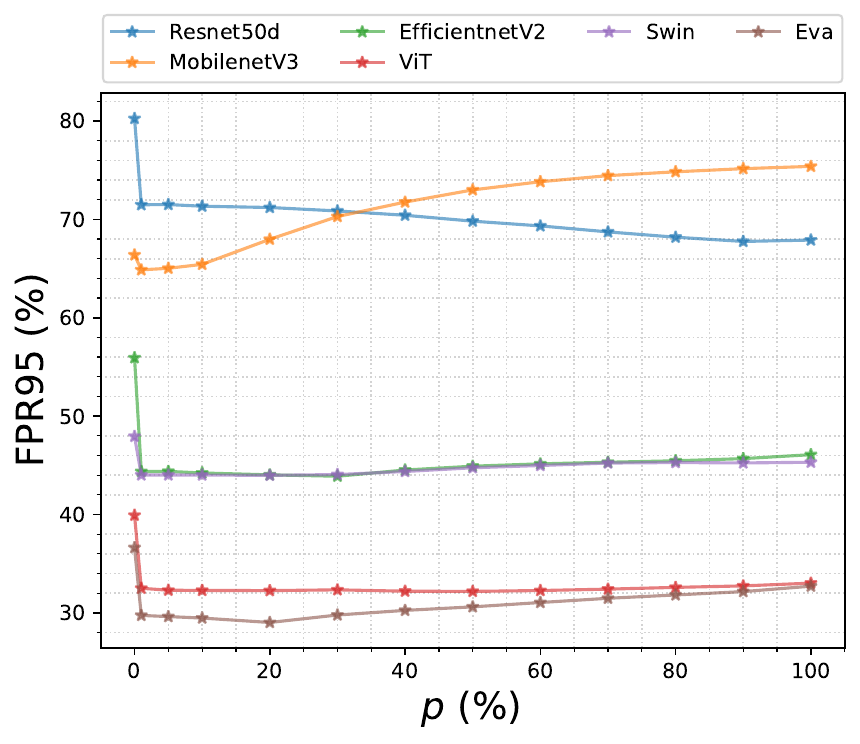}
    \end{minipage}
  \vspace*{-0.6cm}
\end{center}
  \caption{AUROC and FPR95 of ATLI when varying the number of logits used. Left: AUROC; right: FPR95. The x-axis represents the proportion of logits used. The scores appear as the average across all datasets.}
  \label{fig:effect_num_logit}
  \end{figure}
  
\noindent \textbf{Investigation of the number of logits used}
We conducted a test to investigate the effects of the number of top-k logits. The results can be found in Figure \ref{fig:effect_num_logit}. Here, $p$ indicates the top percentage of all logits to be used. $p$=0\% represents MaxLogit, which uses only the top-1 logit ($|M|=0$). Initially, we show that adding only a few top-k logits delivers a major improvement over MaxLogit. Subsequently, as the number of logits used increases, a performance decline can be observed in almost all models.
Therefore, instead of using a single logit or all logits, selectively utilizing a limited set of effective logits leads to improved performance. \\

\begin{figure}[t]
  \centering
  \begin{center}
    \begin{minipage}[b]{0.49\columnwidth}
        \centering
        \includegraphics[width=1\columnwidth]{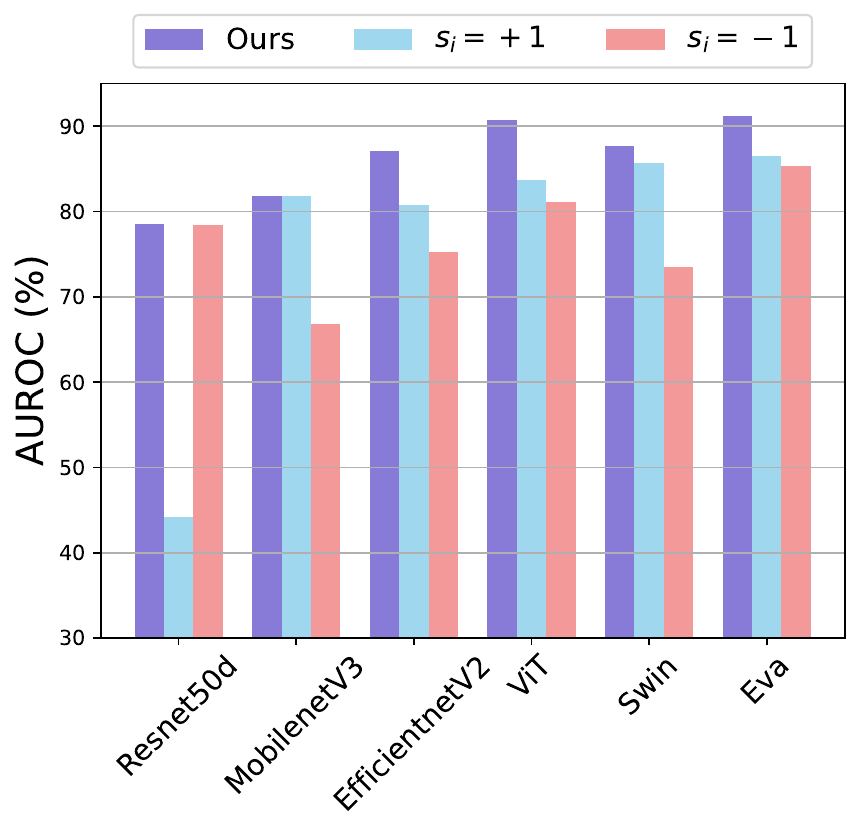}
    \end{minipage}
    \begin{minipage}[b]{0.49\columnwidth}
        \centering
        \includegraphics[width=1\columnwidth]{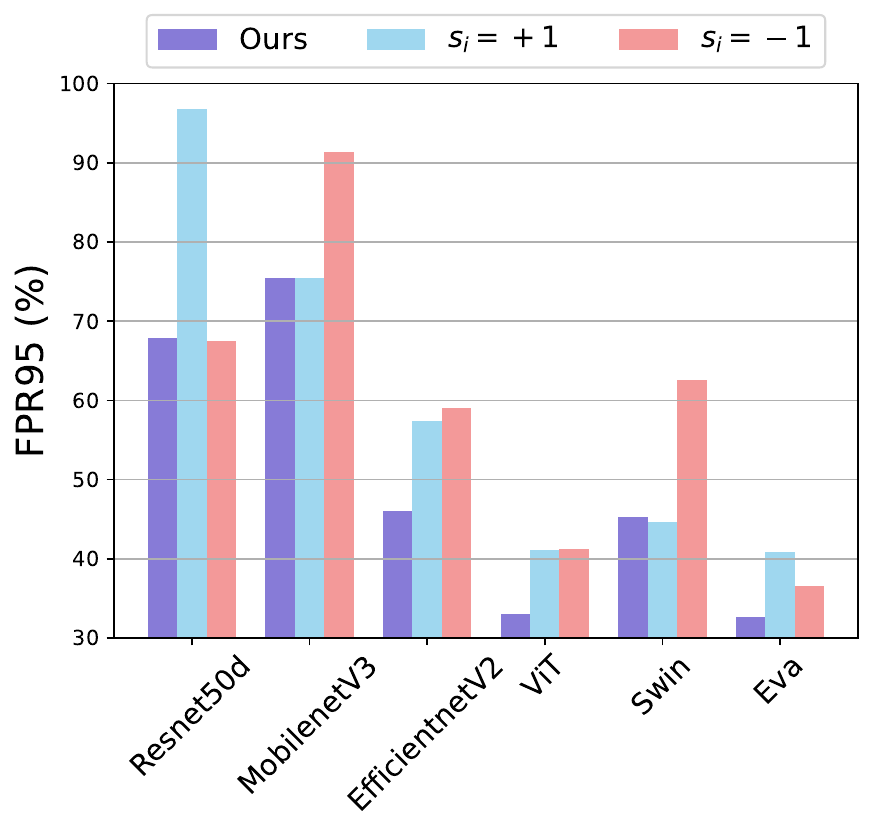}
    \end{minipage}
  \vspace*{-0.6cm}
\end{center}
  \caption{The AUROC and FPR95 of ATLI for each model on changing the sign function in ImageNet-1k benchmark. Here, ATLI uses all logits ($|M|=999$). The purple columns illustrate use of the sign function from Eq. \ref{eq:sign}. Blue columns: $s_k=+1$ and red columns: $s_k=-1$. The left-hand chart shows AUROC, while the right-hand chart shows FPR95. The scores represent the average across all datasets.}
  \label{fig:effect_sign}
  \end{figure}
  
\noindent \textbf{Validity of sign}
We investigated the impact of sign. Figure \ref{fig:effect_sign} shows the AUROC and FPR95 for each model when $p$ is set at 100\%, with the sign obtained from Eq. \ref{eq:sign}, fixed at negative, and fixed at positive.
The figure clearly shows that trends vary significantly between models. For example, ResNet-50d performs better when the sign is negative rather than positive, whereas MobileNetV3 achieves better scores with a positive sign. However, adapting the sign obtained from pseudo-OOD to score function results in an improved score for all models. 
Thus, determining the sign based on Eq. \ref{eq:sign} captures the tendencies of each model, demonstrating robustness across different models.\\

\begin{table*}[t]
  \centering
  \small
  \resizebox{\textwidth}{!}{
  \begin{tabular}{@{}l c@{}lc@{}lc@{}lc@{}lc@{}lc@{}lc@{}l @{}}
    \toprule \\
    \multirow{2}{*}{\textbf{Method}} & \multicolumn{2}{c}{\textbf{ResNet-50d}} & \multicolumn{2}{c}{\textbf{MobilenetV3}} & \multicolumn{2}{c}{\textbf{EffienetNetV2}} & \multicolumn{2}{c}{\textbf{Vit}} & \multicolumn{2}{c}{\textbf{Swin}}  & \multicolumn{2}{c}{\textbf{Eva}} & \multicolumn{2}{c}{\textbf{Average}}\\
    & \scriptsize AUROC\(\uparrow\) & \scriptsize FPR95\(\downarrow\)& \scriptsize AUROC\(\uparrow\) & \scriptsize FPR95\(\downarrow\)& \scriptsize AUROC\(\uparrow\) & \scriptsize FPR95\(\downarrow\)& \scriptsize AUROC\(\uparrow\) & \scriptsize FPR95\(\downarrow\) & \scriptsize AUROC\(\uparrow\) & \scriptsize FPR95\(\downarrow\) & \scriptsize AUROC\(\uparrow\) & \scriptsize FPR95\(\downarrow\) & \scriptsize AUROC\(\uparrow\) & \scriptsize FPR95\(\downarrow\) \\
    \midrule \\
    Mixup &78.89	&72.36	&84.01	&66.49	&87.08	&48.94	&89.9	&35.46	&86.06	&44.85	&92.19	&30.64	&86.36	&49.79 \\
    VOS &51.43	&95.44	&65.89	&93.15	&78.25	&53.09	&90.61	&32.16	&84.89	&53.03	&91.98	&29.1	&77.18	&59.33 \\
    JIGSAW &78.89	&72.34	&79.97	&80.05	&87.83	&44.93	&89	&36.95	&86.27	&50.27	&92.25	&30.28	&85.70	&52.47 \\
    Mixup+VOS &78.68	&71.31	&84.14	&65.40	&87.14	&44.27	&90.60	&32.31	&88.30	&43.97	&92.28	&29.43	&86.86	&47.78 \\
    \bottomrule
  \end{tabular}
  }
  \caption{Performance of different pseudo-OOD. The scores were measured on the ImageNet-1K benchmark and represent the average across all datasets.}
    \label{tab:compare_pseudo}
\end{table*}
\noindent \textbf{Comparison with pseudo-OOD}
In Table \ref{tab:compare_pseudo}, we show the results of using various pseudo OODs. Jigsaw is often used as a pseudo-OOD \citep{featurenorm2023cvpr}. 
Our proposed pseudo-OOD (Mixup+VOS) has the highest AUROC and lowest FPR95. The fact that the score is higher compared to when using Mixup alone or VOS alone indicates that the combination is able to cover a larger portion of the input space.\\

\begin{table*}[t]
  \centering
  \small
  \resizebox{\textwidth}{!}{
  \begin{tabular}{@{}l c@{}lc@{}lc@{}lc@{}lc@{}lc@{}lc@{}l @{}}
    \toprule \\
    \multirow{2}{*}{\textbf{Method}} & \multicolumn{2}{c}{\textbf{ResNet-50d}} & \multicolumn{2}{c}{\textbf{MobilenetV3}} & \multicolumn{2}{c}{\textbf{EffienetNetV2}} & \multicolumn{2}{c}{\textbf{Vit}} & \multicolumn{2}{c}{\textbf{Swin}}  & \multicolumn{2}{c}{\textbf{Eva}} & \multicolumn{2}{c}{\textbf{Average}}\\
    & \scriptsize AUROC\(\uparrow\) & \scriptsize FPR95\(\downarrow\)& \scriptsize AUROC\(\uparrow\) & \scriptsize FPR95\(\downarrow\)& \scriptsize AUROC\(\uparrow\) & \scriptsize FPR95\(\downarrow\)& \scriptsize AUROC\(\uparrow\) & \scriptsize FPR95\(\downarrow\) & \scriptsize AUROC\(\uparrow\) & \scriptsize FPR95\(\downarrow\) & \scriptsize AUROC\(\uparrow\) & \scriptsize FPR95\(\downarrow\) & \scriptsize AUROC\(\uparrow\) & \scriptsize FPR95\(\downarrow\) \\
    \midrule \\
    ATLI (10-80) &79.51	&74.05	&83.04 	&60.84 &87.69	&47.18	&90.91	&33.52	&86.73	&49.39	&87.25	&41.18	&86.75 	&50.20 \\
    ATLI (top-k) &78.68	&71.31	&84.14	&65.40	&87.14	&44.27	&90.60	&32.31	&88.30	&43.97	&92.28	&29.43	&86.86	&47.78 \\
    \bottomrule
  \end{tabular}
  }
  \caption{Performance of adaptively determining top-k logits for each model and Using a consistent top-k for all models. ATLI (10-80) use logit from top-10 to top-80.}
    \label{tab:ablation_topk}
\end{table*}
\noindent \textbf{Validity of model-adaptive top-k}
To evaluate the effectiveness of model-specific top-k selection, we conducted a series of experiments. Table~\ref{tab:ablation_topk} presents the AUROC and FPR95 scores for ATLI(top-k), which adaptively sets the top-k subset $M$ for each model based on pseudo-OOD data, and ATLI(10–80), which uses a fixed $M$ from top-10 to top-80 for all models. This range of top-10 to top-80 was selected based on Figure \ref{fig:auroc_all_topk}, which shows that it yields high scores for most models. Experimental results demonstrate that adaptively determining the top-k logits per model outperforms the use of a fixed value in nearly all cases.
These findings validate the effectiveness of leveraging pseudo-OOD data to adaptively select top-k values for each model. However, for MobileNetV3, the adaptively selected top-k led to a slight drop in performance. This indicates that deriving optimal top-k selections based on pseudo-OOD data is still a challenging problem and highlights the need for further investigation in future work. \\

\begin{table}[t]
  \centering
  \small
  \resizebox{\textwidth}{!}{
  \begin{tabular}{@{}l c@{}lc@{}lc@{}lc@{}lc@{}lc@{}lc@{}l @{}}
    \toprule \\
    \multirow{2}{*}{\textbf{Number of data}} & \multicolumn{2}{c}{\textbf{ResNet-50d}} & \multicolumn{2}{c}{\textbf{MobilenetV3}} & \multicolumn{2}{c}{\textbf{EffienetNetV2}} & \multicolumn{2}{c}{\textbf{Vit}} & \multicolumn{2}{c}{\textbf{Swin}}  & \multicolumn{2}{c}{\textbf{Eva}} & \multicolumn{2}{c}{\textbf{Average}}\\
    & \scriptsize AUROC\(\uparrow\) & \scriptsize FPR95\(\downarrow\)& \scriptsize AUROC\(\uparrow\) & \scriptsize FPR95\(\downarrow\)& \scriptsize AUROC\(\uparrow\) & \scriptsize FPR95\(\downarrow\)& \scriptsize AUROC\(\uparrow\) & \scriptsize FPR95\(\downarrow\) & \scriptsize AUROC\(\uparrow\) & \scriptsize FPR95\(\downarrow\) & \scriptsize AUROC\(\uparrow\) & \scriptsize FPR95\(\downarrow\) & \scriptsize AUROC\(\uparrow\) & \scriptsize FPR95\(\downarrow\) \\
    \midrule \\
    100,000 &78.68	&71.31	&84.14	&65.40	&87.14	&44.27	&90.60	&32.31	&88.30	&43.97	&92.28	&29.43	&86.86	&47.78 \\
    50,000 &78.68	&71.30	&84.13	&65.42	&87.17	&44.28	&90.61	&32.31	&88.32	&43.99	&92.29	&29.46	&86.87	&47.79 \\
    10,000 &78.65	&71.10	&84.13	&65.41	&87.25	&44.28	&90.83	&32.02	&88.41	&44.21	&92.29	&29.50	&86.93	&47.75 \\
    5,000 &78.70	&71.39	&84.13	&65.40	&87.47	&44.35	&90.87	&32.01	&88.34	&44.58	&92.27	&29.45	&86.96	&47.86 \\
    1,000 &78.51	&70.11	&84.16	&65.36	&87.72	&44.51	&91.09	&32.29	&87.90	&45.89	&92.19	&30.40	&86.93	&48.09 \\
    \bottomrule
  \end{tabular}
  }
  \caption{Performance of ATLI when varying the number of samples used for parameter estimation.}
    \label{tab:ablation_numdata}
\end{table}

\noindent \textbf{Number of Samples for ATLI Setup}
In real-world applications, it is often impractical to access large numbers of in-distribution samples for post-hoc calibration. To simulate such settings, we varied the number of ID samples used to estimate the parameters required by ATLI, such as $M$ and sign values. Specifically, we randomly sampled 100,000, 50,000, 10,000, 5,000, and 1,000 training images from ImageNet-1K for this setup procedure as shown in Table \ref{tab:ablation_numdata}. We observed that the performance of ATLI remained stable even with as few as 1,000 samples. These results indicate that ATLI is data-efficient and well suited for real-world applications.

\section{Conclusion}
We proposed ATLI, an adaptive method for OOD detection that integrates the maximum logit with a selected subset of top-k logits based on pseudo-OOD samples. Unlike existing methods that use only the maximum logit or all logits, ATLI focuses on model-specific informative logits. Experiments on ImageNet-1K benchmark show that ATLI consistently improves AUROC and FPR95 over strong baselines, including TRIM and ViM. Our findings emphasize the necessity of selectively utilizing model-specific logits, as opposed to fixed or exhaustive strategies, to achieve robust and efficient OOD detection.





\bibliography{reference}






\end{document}